%% file: main.tex
\documentclass[runningheads]{llncs}

\usepackage{eccv} 

\usepackage{eccvabbrv}

\usepackage{graphicx}
\usepackage{booktabs}

\usepackage[accsupp]{axessibility}  

\usepackage{marvosym}
\usepackage{amsmath}
\usepackage{bbm}
\usepackage{graphicx}
\usepackage{booktabs}
\usepackage{url}
\usepackage{wrapfig}
\usepackage{algorithm}
\usepackage{algpseudocode}

\usepackage{indentfirst}
\usepackage{footnote}
\usepackage{enumerate}

\usepackage{colortbl}
\definecolor{mygray}{gray}{0.97}
\definecolor{darkgreen}{RGB}{34,200,34}
\definecolor{pp}{RGB}{243,235,255}
\newcommand{\green}[1]{\textcolor{darkgreen}{#1}}
\newcommand{\gbf}[1]{\green{\bf{#1}}}

\usepackage[pagebackref,breaklinks,colorlinks,citecolor=eccvblue]{hyperref}

\usepackage{orcidlink}

\begin{document}

\title{SUMix: Mixup with Semantic and Uncertain Information} 


\titlerunning{SUMix} 

\author{Huafeng Qin\inst{1,2,5, \star, {\textrm{\Letter}}}\orcidlink{0000-0003-4911-0393},
Xin Jin\inst{1,2,\star}\orcidlink{0009-0005-0983-6853},
Hongyu Zhu\inst{1,2,\star}\orcidlink{0009-0000-5993-4666},
Hongchao Liao\inst{1,2},\\
Mounîm A. El-Yacoubi\inst{3}\orcidlink{0000-0002-7383-0588},
Xinbo Gao\inst{4}
}
\authorrunning{Qin. et al.}

\institute{Chongqing Technology and Business University
, Chongqing, 400067, China \and
National Research base of Intelligent Manufacturing Service, Chongqing, China \and
Telecom SudParis, Institut Polytechnique de Paris, Paris, 75020, France \and
Chongqing University of Posts and Telecommunications, Chongqing, 400065, China \and
Chongqing Micro-Vein Intelligent Technology Co. \\
\email{\{2013009, jinxin4, zhuhongyu\}@ctbu.edu.cn}\\
$\star$ Equal contribution, \textrm{\Letter} Corresponding author.
}

\maketitle

\input{0_Abstract}
\input{1_Introduction}
\input{2_Related_work}
\input{3_Preliminaries}
\input{4_SUMix}
\input{5_Experiments}
\input{6_Conclusion}
\section*{Acknowledgement}
This work is supported in part by the Scientific Innovation 2030 Major Project for New Generation of AI under Grant 2020AAA0107300, in part by the National Natural Science Foundation of China (Grant No. 61976030), in part by Chongqing Natural Science Foundation (Grant No. 2024NSCQ-MSX4893), in part by Project of CQ CSTC (Grant No. cstc2018jcyjAX 0057), in part by Innovative Research Projects for Postgraduate Students of Chongqing Business University (No. yjscxx2024-284-53, No. yiscxx2024-284-231, No. yjscxx2024-284-234).

%
%
\bibliographystyle{splncs04}
\bibliography{main}

\clearpage
\input{Appendix}

\end{document}

%% file: 0_Abstract.tex
\begin{abstract}
Mixup data augmentation approaches have been applied for various tasks of deep learning to improve the generalization ability of deep neural networks. Some existing approaches CutMix, SaliencyMix, \emph{etc.} randomly replace a patch in one image with patches from another to generate the mixed image. Similarly, the corresponding labels are linearly combined by a fixed ratio $\lambda$ by $l$. The objects in two images may be overlapped during the mixing process, so some semantic information is corrupted in the mixed samples. In this case, the mixed image does not match the mixed label information.  Besides, such a label may mislead the deep learning model training, which results in poor performance.  To solve this problem,  we proposed a novel approach named SUMix to learn the mixing ratio as well as the uncertainty for the mixed samples during the training process. First, we design a learnable similarity function to compute an accurate mix ratio. Second,  an approach is investigated as a regularized term to model the uncertainty of the mixed samples. We conduct experiments on five image benchmarks, and extensive experimental results imply that our method is capable of improving the performance of classifiers with different cutting-based mixup approaches. The source code is available at 
\href{https://github.com/JinXins/SUMix}{https://github.com/JinXins/SUMix}.
\keywords{mixup, data augmentation, image classification}
\end{abstract}

%% file: 1_Introduction.tex
\section{Introduction}
\label{sec:intro}

Deep learning has emerged as a viable alternative to traditional machine learning in various tasks, \emph{e.g.} computer vision \cite{2020YOLOv4, li2022moganet, krizhevsky2012imagenet}, natural language processing \cite{vaswani2017attention, devlin2018bert, icml2024CHELA}, and video applications \cite{iccv2019SlowFast, ijcai2020tlpg, tan2023temporal}, because of its robust feature representation capacity. However, deep learning requires a lot of data to train thousands of network parameters. Despite the self-supervised learning paradigm \cite{cvpr2020moco, icml2023a2mim, li2024genurl, iclr2024semireward} enables task agnostic pre-training without annotations, the labeled data of specific scenarios is still required and generally limited, where deep neural networks (DNNs) are prone to overfitting~\cite{bishop2006pattern}. Therefore, the representation capacity is not effectively exploited, resulting in poor discriminative performances and generalization abilities.

\begin{figure}[t]
    \centering
    \captionsetup{type=figure}
    \includegraphics[scale=0.28]{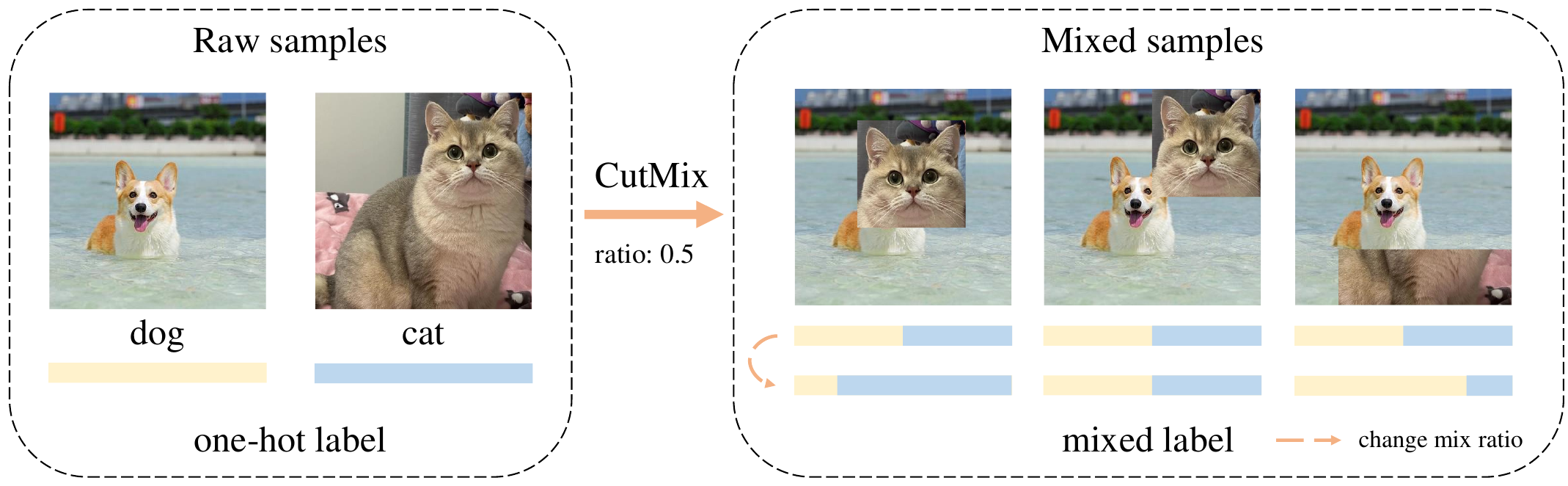}
    \caption{The diagram shows different cases of raw samples that underwent the CutMix with a mixing ratio of 0.5 to obtain mixed samples. $Left:$ the raw samples and their one-hot labels; $Right:$ labels of the mixes obtained using a redefined mixing ratio $\widetilde{\lambda}$ for different cases.}
    \label{fig:1}
\end{figure}

Data Augmentation (DA) methods as a solution have been proposed to prevent DNNs from over-fitting. Existing methods transform original data to generate new samples for data augmentation. For example, in computer vision tasks, classical techniques such as rotating, scaling, and panning have been used to generate augmented samples for model training. Similarly, in natural language processing tasks, methods such as proximate word substitution and randomly misspelling words are employed to produce new data. These data augmentation methods have proven to be effective in improving the performance of network models. They allow the models to learn a larger variety of samples based on augmentation data. Currently, data augmentation methods can be categorized into three classes. 
(\romannumeral1) The contrast combination methods~\cite{cubuk2018autoaugment, lim2019fastda, cubuk2020randaugment, hataya2020fasterda, li2020differentiableaa, suzuki2022teachaug} employs affine matrix transformations and color enhancement for data augmentation. However, the resulting images are significantly different from the original ones. (\romannumeral2) Mixup method produces virtual mixup examples via a simple convex combination of pairs of examples \cite{zhang2018mixup, verma2019manifold, qin2023adautomix}. (\romannumeral3) The generative approaches, \emph{e.g.} Generative Adversarial Networks(GANs) \cite{goodfellow2020generative, yang2022cgan}, generate fake samples by adversarial learning. Overall, these data augmentation methods effectively alleviate the overfitting problem and improve the performance of deep learning models on limited training data.

\begin{figure}[t]
    \centering
    \includegraphics[scale=0.32]{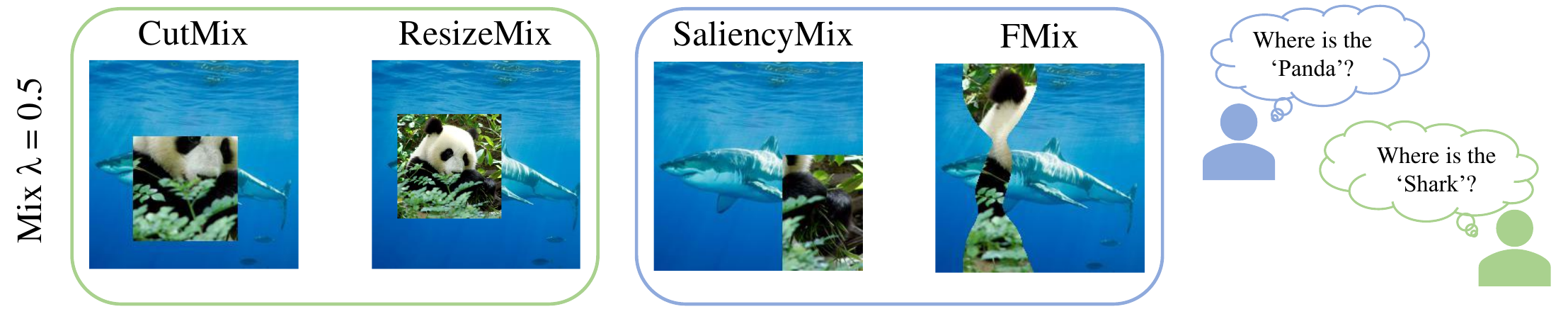}
    \caption{The figure shows some hand-crafted mixup methods with \emph{``Label MisMatch"} problem. In CutMix and ResizeMix, the \textbf{\emph{``Shark"}} is mostly obscured, while in SaliencyMix and FMix, it is difficult to notice the \textbf{\emph{``Panda"}}.}
    \label{fig:3}
\end{figure}

Mixup \cite{zhang2018mixup} method as a recent DA model produces virtual samples and labels by linearly interpolating the sample images as well as their corresponding labels, and such a simple method has proven to be effective in improving the representation capacity of the deep neural networks. In recent years, various mixup methods have been proposed for data augmentation. For example, CutMix~\cite{yun2019cutmix} shifts the numerical mixing to spatial, where the mixing coefficient $\lambda$ determines the size of the area of the cut and introduces the idea of Cutout~\cite{devries2017improved}, which cuts and mixes two samples, as shown in Figure~\ref{fig:1}. This approach not only helps the network to generalize better and improve object localization but also enables continuous performance improvement in tasks such as Pascal VOC object detection~\cite{ren2015fasterrcnn} and MS-COCO image captioning~\cite{vinyals2015show}. In addition, CutMix improves the robustness of the model, making it perform better in the face of input corruption and cross-distribution detection. However, the CutMix method randomly selects the cropping region, which results in the background region of the target sample being clipped to the feature region of the source sample, and in the extreme case, the source sample will be completely occluded, resulting in an augmented sample that has no feature information at all, a phenomenon known as ``\emph{Label MisMatch}" shown in Figure.\ref{fig:3}. Since the one-hot label of the sample is unchanged, Decoupled Mix~\cite{liu2022decoupledmix} found that the Mixup method relies heavily on the loss function, forcing the prediction of the network model to obey two higher peaks, which results in an inaccurate gradient result due to the label mismatch problem in the calculation of the loss function. This leads to an inaccurate gradient result when calculating the loss function, which affects the training of the network model.

\begin{figure}[b]
    \centering
    \includegraphics[scale=0.23]{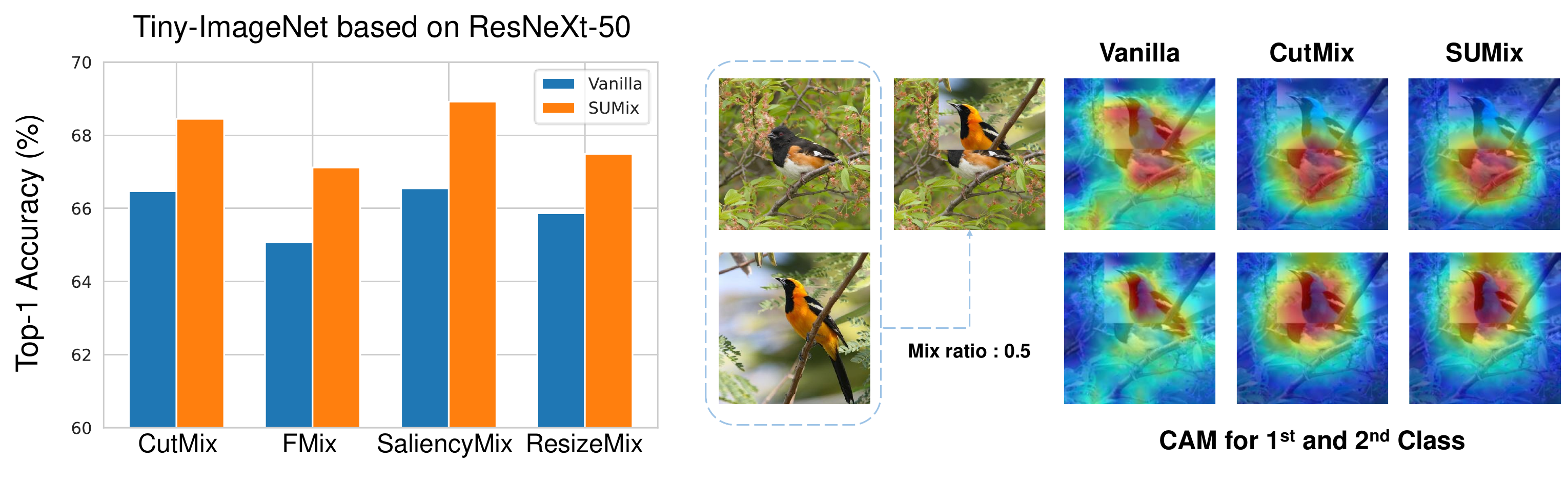}
    \vspace{-0.5em}
    \caption{The results illustrate the application of SUMix. \emph{Left:} Top-1 Accuracy improvement from the mixup approaches with SUMix; \emph{Right:} Comparison of Vanilla method, CutMix, and SUMix for CAM visualization.}
    \label{fig:2}
\end{figure}

To improve the ``\emph{Label MisMatch}" problem, we propose \textbf{SUMix}, which consists of a mix ratio learning module and an uncertain estimation module. 
The former focuses on computing the proportion of two images and the latter aims to learn the uncertainty information of mixed samples. Firstly, we design a function to compute the semantic distance between the mixed and original samples to determine the ratio $\lambda$. Secondly, we present a method to learn the uncertainty of the mixed. This adapted feature vector effectively mitigates issues related to computing the loss function caused by discrepancies in semantic and uncertainty aspects. Extensive experiments on five image classification datasets verify that our proposed SUMix can remarkably improve performances of existing mixup augmentations (as shown in Figure~\ref{fig:2}) in a plug-and-play manner while achieving better robustness.
Our main contributions are as follows:
\begin{enumerate}[(a)]
\setlength\topsep{0.0em}
\setlength\itemsep{0.1em}
\setlength\leftmargin{1.0em}
    \item We propose a learnable metric to compute the mixed ratio by similarity between the mixed samples and the original samples.
    \item  We further consider the uncertainty and semantic information of the mixed samples and recalculate a reasonable feature vector, providing an additional regularization loss for model training.
    \item Our approach helps popular Cutting-based mixup methods to improve classification tasks without spending too excessive extra time overhead.
\end{enumerate}

%% file: 2_Related_work.tex
\section{Related Work}
\label{sec: related work}

\paragraph{\textbf{Sample-based Mixup}} Sample-based Mixup approaches~\cite{lee2020smoothmix, cvpr2022pixmix, hendrycks2019augmix, huang2020snapmix, yang2022recursivemix} focus on obtaining more reliable mixed samples or hidden space feature maps. Usually combined with inserting prior knowledge or saliency information~\cite{selvaraju2017gradcam}. CutMix~\cite{yun2019cutmix} obtains the mixed samples by randomly cutting and pasting between the two samples. Due to the large randomness of CutMix cropping, ResizeMix~\cite{qin2020resizemix} replaces the cut operation with the resize operation. There are a series of works such as FMix~\cite{harris2020fmix}, GridMix~\cite{Baek2021gridmix} \emph{etc}, which use similar methods. To extract the saliency information within the samples during the training of the model, some work such as SaliencyMix~\cite{uddin2020saliencymix}, Attentive-CutMix~\cite{icassp2020attentive} \emph{etc}, use a saliency extraction module to mix the samples in a guidedness. PuzzleMix~\cite{kim2020puzzle} extracts the saliency regions from the source and target samples respectively through the saliency capability learned from the model and then extracts the saliency regions according to the designed ``\emph{optimal transmission}". Co-Mixup~\cite{kim2020co} further improves PuzzleMix by increasing the number of mixed samples from 2 to 3. AutoMix~\cite{liu2022automix} and SAMix~\cite{li2021samix} analyze the advantages and disadvantages of the hand-crafted and saliency method and divide the mixing task into two sub-tasks. AdAutoMix~\cite{qin2023adautomix} further improves AutoMix by proposing adversarial training to train the generator and obtain ``\emph{hard mixed sample}" to train the model. Manifold Mixup~\cite{verma2019manifold} and AlignMix~\cite{2021alignmix} moved the mixup method to hidden space by mixing the feature information of the samples.

\paragraph{\textbf{Label-based Mixup}} Label-based Mixup approaches aim to solve the problems of 1) time overhead and 2) sample-label mismatch. These methods usually use some known information to recalculate the mix ratio of labels. Saliency Grafting~\cite{park2021saliencygraft} uses the area of saliency features of each sample in a mixed sample to reconstruct a new $\lambda$. LUMix~\cite{sun2022lumix} differs from previous ways to address the label noise introduced by the mixup process instead of modeling the label noise. Methods such as TransMix~\cite{chen2022transmix} and TokenMix~\cite{liu2022tokenmix} use attention scores to mix the samples to compute $\lambda$ based on the area of the mask. Mixpro~\cite{zhao2023mixpro} obtains a new $\lambda$ by combining the attention score with the masked area and obtains a stable output with a factor $\alpha$. Other methods like SMMix~\cite{chen2023smmix}, Token Labeling~\cite{jiang2021tokenlabeling} \emph{etc}, are based on attention score and masking area to get a mix ratio.

\paragraph{\textbf{Uncertainty Modeling}} The presence of naturally occurring uncertainties such as occlusion~\cite{kendall2017uncertainties} and blurring~\cite{shaw2002signal} in the images tends to influence the training of network models. Several works~\cite{chang2020unface, shi2019probabilistic, zhang2021point} have attempted to model uncertainty in face recognition, age estimation, and point cloud segmentation, respectively, to obtain better robustness and generalization. In addition, it is also able to deal with Out-distribution problems. \cite{guenais2020bacoun} proposed a Bayesian classifier for obtaining Out-of-distribution uncertainty. \cite{oh2018modeling} defines a Gaussian distribution to represent the samples and uses Monte Carlo sampling to adopt multiple points from the Gaussian distribution for optimizing the metric. \cite{wang2023introspective} uses an introspective modeling approach that bypasses distribution optimization to further improve the robustness of the model.

%% file: 3_Preliminaries.tex
\section{Preliminaries}
\label{sec: prelim}

\paragraph{\textbf{MixUp Classification.}} Let us define an $X \in \mathbb{X}$ as the input data and its corresponding $Y \in \mathbb{Y}$ as the label of the input data. For the vanilla mixup, we construct a mixed sample $\widetilde{X}$ and a mixed label $\widetilde{Y}$ from Eq.(\ref{eq:1}).
\begin{equation}
\begin{aligned}
    \widetilde{X}=\lambda\ast X_a+(1-\lambda)\ast X_b, \\
    \widetilde{Y}=\lambda\ast Y_a+(1-\lambda)\ast Y_b,
    \label{eq:1}
\end{aligned}
\end{equation}
where $\lambda$ is a mix ratio obtained from a Beta distribution, $(X_a, X_b)\in\widetilde{X}$, and $(Y_a, Y_b)\in\widetilde{Y}$ are the same batch of data, Mixup obtains the mixed samples by pixel interpolating with a $\lambda$, CutMix randomly obtains a two-valued rectangular mask $\mathbbm{1}$, where $\mathbbm{1}=\lambda=\frac{r_wr_h}{WH}$, W, H are the width and height of the data and $r_w, r_h$ are the width and height of the rectangular mask. We get the mixed samples $\widetilde{X}=\mathbbm{1}\odot X_a + (1-\mathbbm{1})\odot X_b$.

\paragraph{\textbf{The Devil Lies in Label MisMatch.}} Mixup contains two different labels, so the Mixed Cross Entropy (MCE) is used in calculating the loss function as shown in Eq.(\ref{eq:2}).
\begin{equation}
\begin{aligned}
    \mathcal{L}_{MCE} = \ell_{CE}(f_\theta(\widetilde{X}), Y_a)\ast\lambda + \ell_{CE}(f_\theta(\widetilde{X}), Y_b)\ast(1-\lambda),
    \label{eq:2}
\end{aligned}
\end{equation}
where $f_\theta\left(\cdot\right)$ denotes the network model, and $\ell\left(\cdot\right)$ denotes the loss function, $\ell_{CE}$ was cross-entropy loss, which minimizes the loss during the training process. However, since some hand-crafted mixup methods are random, \emph{e.g.} CutMix, ResizeMix, but the labels of the mixed samples are fixed $\widetilde{Y}$, then in the case shown in Figure.\ref{fig:3}, the features of a certain classification are occluded and reduced or do not exist in the mixed samples, resulting in the feature information not matching the labels. The phenomenon is called ``Label MisMatch". In this case, the proportion of labels should be reduced or removed to get an accurate result when calculating the loss function.

%% file: 4_SUMix.tex
\section{SUMix}
\label{sec: sumix}
\begin{figure}[t]
    \centering
    \includegraphics[scale=0.28]{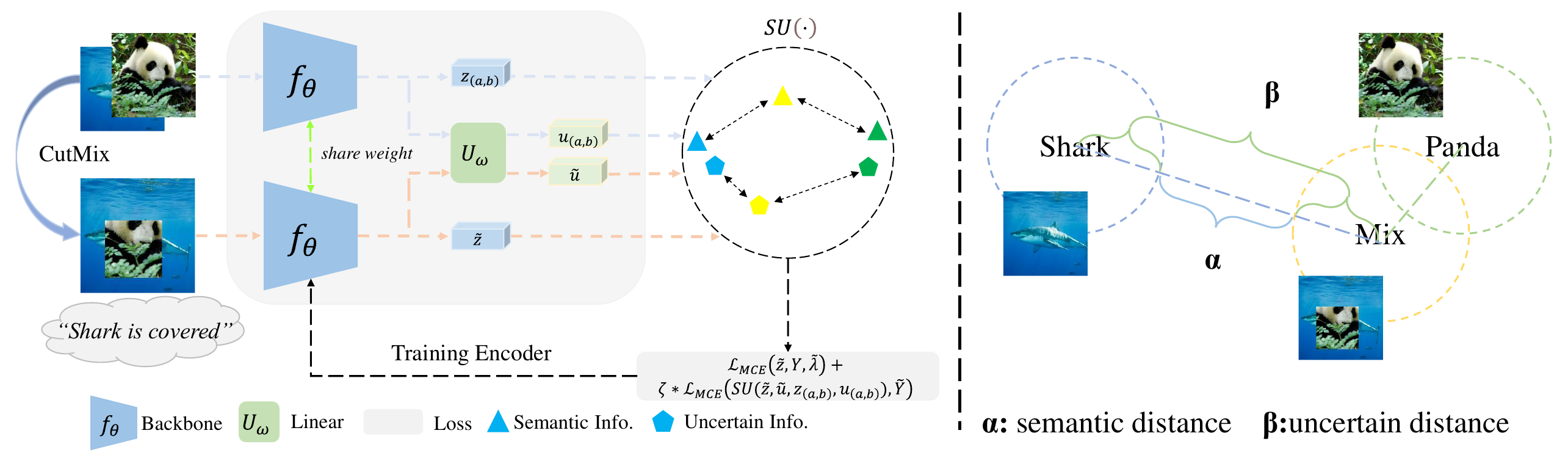}
    \caption{The left diagram represents the training process of SUMix. The raw and mixed samples are encoded through the model to obtain the semantic and uncertainty information, and the new mix ratio and loss function for regularization are obtained. The right diagram is the spatial distance between the raw and mixed samples.}
    \label{fig:4}
\end{figure}

In this section, we focus on detailing SUMix, 1) the uncertainty classifier, 2) how to obtain the mix ratio $\lambda$, and 3) the regularization function in the case of uncertainty and semantic information. Firstly, we show how to use deep learning classifiers for loss function computation and network model training, secondly, how to redefine the mix ratio $\lambda$ based on semantic information, and lastly, how to consider uncertainty and regularization. The pipeline of SUMix is shown in Fig. \ref{fig:4}.

\subsection{Uncertainty Classifier}
Assuming $\mathbb{S}=\left\{ x_s|s=1,2,\cdots ,S\right\}$ is a training set and $\mathbb{S}$ is the number of training images. We take $N$ samples from it as a sample set, where $\mathbb{X}\in \mathbb{R}^{N\times 3\times H\times W}$, $\mathbb{Y}\in \mathbb{R}^{N\times K}$ are the labels and $K$ is the number of classes. For the original computation of the mixup loss function, we use a deep neural network $f_\theta\left(\cdot\right)$ with a trainable parameter that maps the sample $x_i$ to the labels $y_i$. Unlike the vanilla mixup loss computation in Eq.(\ref{eq:2}), SUMix adds a regularized loss function as shown in Eq.(\ref{eq:3}): 
\begin{equation}
\begin{aligned}
    \mathcal{L}_{su} = \frac{1}{N}\left(\sum_{i=0}^{N}\mathcal{L}_{MCE}(f_\theta( \widetilde{X}_i), Y, \widetilde{\lambda}_i ) \right)+ \zeta \ast \mathcal{L}_{MCE}\left(SU( f_\theta ( \widetilde{X}), U_\omega ),\widetilde{Y}\right),
    \label{eq:3}
\end{aligned}
\end{equation}
where $\widetilde{ \lambda }_i$ are the redefined lambda, $SU(\cdot)$ denotes a function to obtain feature, and $\zeta$ is a hyperparameter used to weight the regulated loss function, set to [\emph{0.1}, \emph{0.2}, \emph{0.5}, \emph{1.0}]. $U_\omega$ denotes a function as a measure of uncertainty.
Finally, the parameters of the deep neural network are updated continuously to minimize $\mathcal{L}_{su}$.

\subsection{Mix ratio Learning}
Mixup approaches always depend on a parameter taken from the $Beta (\alpha, \alpha)$ distribution, which we call the mix ratio $\lambda$, to balance the mix ratio of the two samples. Lots of mixup approaches use $\lambda$ to augment the mixup sample set $\mathbb{X}$ after obtaining it, and notably, this $\lambda$ serves the whole sample set. For some saliency-based or automatic-based mixups, \emph{e.g.} PuzzleMix~\cite{kim2020puzzle}, AutoMix~\cite{liu2022automix}, it is possible to push the mixing method to obtain the best possible mixed samples according to the $\lambda$ ratio, balancing the sample ratio with the $\lambda$ as much as possible. However, some hand-crafted approaches with randomness cannot accurately match the mix ratio. The problem we want to solve is how to match each mixup sample with its own $\lambda$.

For a mixed sample $\widetilde{x}=\left[\widetilde{X}\right]^N_i\in R^{3\times H\times W}$, semantic information links need to be established with the original sample pair $\left(x_a,x_b\right)$. They are fed into a deep neural network $f_\theta(\cdot)$, \emph{e.g}. ResNet18~\cite{he2016deep}. feature vectors $\widetilde{z}$ and $\left(z_a,z_b\right)$ are obtained, and the feature vectors are normalized to modify their similarity according to Eq.(\ref{eq:4}):
\begin{equation}
\begin{aligned}
    \widetilde{\lambda}_a = \frac{\lambda \ast \mathrm{e}^{-\parallel(\sigma(\widetilde{z}-z_a)\parallel_2} }{\lambda \ast \mathrm{e}^{-\parallel(\sigma(\widetilde{z}-z_a)\parallel_2} + (1-\lambda) \ast \mathrm{e}^{-\parallel(\sigma(\widetilde{z}-z_b)\parallel_2} },
    \label{eq:4}
\end{aligned}
\end{equation}
where $\sigma(\cdot)$ is the softmax function used to normalize the vectors, and ${\parallel \cdot \parallel}_2$ is the l2 norm, which imposes a limit on the distance between the two. When the difference in semantic information between $\widetilde{x}$ and $x_a$ is large, which indicates that the mixed sample has a small amount of the feature information of the raw sample, ${\widetilde{\lambda}}_a$ tends to 0. Similarly, when the difference in semantic information between the two is small, ${\widetilde{\lambda}}_a$ tends to 1. $\widetilde{\lambda}_b = 1 -{\widetilde{\lambda}}_a$.
Thus, we have calculated the mixed sample set with its own matching $\lambda$. a more sensible loss is computed according to Eq.(\ref{eq:3}).

\subsection{Uncertainty Estimation}
Deep learning tasks involve two types of uncertainty: Epistemic uncertainty and Aleatoric uncertainty. Epistemic uncertainty arises from the model's inherent uncertainty due to insufficient training data, resulting in low confidence in unknown data. In contrast, Aleatoric uncertainty arises from errors in the data itself, which can introduce bias into the dataset. The greater the bias, the greater the Aleatoric uncertainty. In the case of the mixup task, the error in the data occurs when the mixup method is stochastic. Our goal is to address this error and enable the network to adapt accordingly.

Wang et al.~\cite{wang2023introspective} propose an introspective metric that models a new vector for classification based on the semantic information distance between two images and their uncertainty distance. The mixup methods differ from this as they typically incorporate feature information from two samples. To begin, we encode the mixed sample and the raw sample using an encoder to obtain the feature vectors $\widetilde{z}$ and $z$, respectively. Next, we calculate the uncertainty by passing $\widetilde{z}$ and $z$ through $U_\omega$, by Eq.(\ref{eq:5}).
\begin{equation}
\begin{aligned}
    u = \parallel \sigma(MLP(z)) \parallel_2, \\
    \widetilde{u} = \parallel \sigma(MLP(\widetilde{z})) \parallel_2,
    \label{eq:5}
\end{aligned}
\end{equation}
where is a linear layer with a softmax function and l2 norm. We define $\beta = \widetilde{u} + u$ as the uncertainty of mixed samples and raw samples.

After capturing the uncertainty and semantic information, as demonstrated in the right part of Figure.\ref{fig:4}, Wang et al.~\cite{wang2023introspective} argue that in the feature space, the smaller the radius of a sample, the more aggregated the class is, and the smaller the uncertainty. By combining the semantic information, we reconstruct the mixed samples $\widetilde{X}$ and the raw samples $X$ to obtain $Z_{su}$ according to Eq.(\ref{eq:6}).
\begin{equation}
\begin{aligned}
    Z_{su} = \mathrm{e}^{-(\beta + \parallel\sigma(\widetilde{z} - z)\parallel_2)}
    \label{eq:6}
\end{aligned}
\end{equation}

When there is significant uncertainty or semantic information has a large difference, $Z_{su}$ receives a small gradient; conversely, when there is little uncertainty and small differences in semantic information, $Z_{su}$ receives a normal gradient that tends to 1. This approach is used as a method of regularization.

%% file: 5_Experiments.tex
\section{Experiments}
\label{sec: experiments}
To verify the performance of SUMix, we conducted a large number of experiments on six datasets, CIFAR100~\cite{krizhevsky2009learning}, Tiny-ImageNet~\cite{2017tinyimagenet}, ImageNet-1K~\cite{krizhevsky2012imagenet}, CUB-200~\cite{wah2011caltech}, FGVC-Aircrafts~\cite{maji2013fine}, and CIFAR100-C. To verify the effectiveness of the improved ``Label MisMatching" method, we mainly compared it with some patch-wise mixup methods based on hand-crafted, \emph{i.e.}, CutMix, FMix, ResizeMix, and SaliencyMix. We have implemented our algorithm on the open-source library OpenMixup~\cite{2022openmixup}, and some parameters are designed according to OpenMixup. The classification results are the median of the top-1 test accuracy from the last 10 training epochs, but on ImageNet-1K, we choose the last 5 training epochs. To facilitate comparison, our results are highlighted in \textbf{bold}.

\subsection{Datasets}
We choose the three most classical datasets and two fine-grained datasets in the classification task. CIFAR100 has a total of 50,000 images for the training set and 10,000 images for the testing set, containing 100 different classes, each with a 32$\times$32 resolution; Tiny-ImageNet contains 10,000 images for training and 10,000 images for testing, each with a 64$\times$64 resolution and a total of 200 classes; The ImageNet-1K dataset contains 1.2 million images for training and 50,000 images for validation; CUB200 dataset contains images of 200 different bird species, a total of 11,788 images; the FGVC-Aircrafts dataset contains images of 100 different types of aircraft, a total of 10,000 images.

\subsection{Implementation Details}
\paragraph{\textbf{Hyper-parameter of $\alpha$.}} In the experimental results we report, the hyper-parameter $\alpha$ of the mixup variants, which sampling from the $Beta(\alpha, \alpha)$ distribution used by the model based on the CNN architecture and the model based on the ViT architecture are different. For CutMix, the CNN-based models, \emph{e.g.} ResNet18, ResNeXt50, and Wide-ResNet28 set as \emph{0.2}, and ViT-based models, DeiT and Swin-Transformer set as \emph{2.0}. For FMix, CNN-based set \emph{0.2}, and ViT-based set \emph{1.0}. For PuzzleMix, the CNN-based set is \emph{1.0}, and the ViT-based set is \emph{2.0}. In CNN-based and Vit-based architectures alike, SaliencyMix, ResizeMix, and AutoMix exhibit identical alpha values, which are \emph{0.2}, \emph{1.0}, and \emph{2.0}, respectively.

\paragraph{\textbf{Hyper-parameter of $\zeta$.}} In SUMix, we set the hyper-parameters $\zeta$ of $\mathcal{L}_{su}$ as follows: On the CIFAR100 dataset, we report five different models, the $\zeta$ setting shown in Appendix A.1. On Tiny-ImageNet, we set the $\zeta$=\emph{1.0} for all mixup methods based on ResNet18 and ResNeXt50. Similarly with ImageNet-1K, we set the $\zeta$=\emph{0.5} based on ResNet18. The settings we reported for the fine-grained datasets are also shown in Appendix A.1.

\input{tabs/exp_generic}

\subsection{Image Classification}
\paragraph{\textbf{Generic classification.}} We initially train ResNet18~\cite{he2016deep} and ResNeXt50~\cite{xie2017aggregated} using the CIFAR100 dataset for a total of 800 epochs. The experimental parameters are set as in OpenMixup~\cite{2022openmixup}: a basic learning rate of 0.1, SGD~\cite{loshchilov2016sgdr} optimizer with momentum of 0.9, weight decay of 0.0001, and batch size of 100 for each iteration. Wide-ResNet~\cite{bmvc2016wrn} using the CIFAR100 dataset for a total of 400 epochs, learning rate of 0.03, SGD optimizer with momentum of 0.9, weight decay of 0.001. For the Tiny-ImageNet dataset, we use the same experimental parameters as the CIFAR100 experiment, except the base learning rate to 0.2 and training for 400 epochs. As for ImageNet-1K, we follow the Pytorch-style settings for training ResNet18 of 100 epochs. For the CIFAR100 dataset, we use basic data augmentation, including \texttt{RandomFlip} and \texttt{RandomCrop} with 4 pixels padding, and for the Tiny-ImageNet and ImageNet-1K datasets, we also use the \texttt{RandomFlip} and \texttt{RandomResizedCrop} as the default data augmentation operations.

Table \ref{tab:gener} shows our five comparisons on three different datasets, where the use of SUMix brings some improvement in the classification accuracy of CutMix, FMix, SaliencyMix, and ResizeMix. For the CIFAR100 dataset, on ResNet and its variants ResNeXt50, W-ResNet28-8, SUMix improves their classification results by an average of \gbf{0.82}\%, \gbf{1.07}\%, and \gbf{0.12}\%, respectively; for the Tiny-ImageNet dataset, on ResNet18 and ResNeXt50, the average improvements of \gbf{0.81}\% and \gbf{2.07}\%; for the ImageNet-1K dataset, on ResNet18, the average improvement is \gbf{0.47}\%.

\paragraph{\textbf{Fine-grained classification.}} We load the official PyTorch pre-trained
models on ImageNet-1k as initialization and fine-tuning pre-trained ResNet18 and ResNeXt50 using SGD optimizer with a learning rate of 0.001, a momentum of 0.9, weight decay of 0.0005 on CUB200, FGVC-Aircrafts, total training 200 epochs.
\input{tabs/exp_fine_grained}
\begin{figure*}[t]
    \begin{minipage}{0.42\linewidth}
        \input{tabs/vit_cls}
    \end{minipage}
    \begin{minipage}{0.57\linewidth}
        \input{tabs/saliency_mixup_cls}
    \end{minipage}
\end{figure*}
Table \ref{tab: fine} presents the comparison results for the bird and aircraft datasets. SUMix improves the classification results for the CUB200 dataset by an average of \gbf{0.93}\% and \gbf{0.10}\% at ResNet18 and ResNeXt50, respectively. The FGVC-Aircrafts dataset improves by an average of \gbf{0.82}\% and \gbf{0.67}\% at ResNet18 and ResNeXt50.

\paragraph{\textbf{Classification based on ViTs.}} To further show the performance of SUMix, we perform classification experiments using the ViT-based model DeiT~\cite{icml2021deit} and Swin-Transformer~\cite{iccv2021swin} on the CIFAR100 dataset, and the results are shown in Tab \ref{tab: vit}. We resized the size to 224 $\times$ 224, and in the training set, we trained DeiT and Swin-Transformer of 200 epochs by AdamW~\cite{iclr2019AdamW} optimizer with a batch size of 100 on CIFAR100. The basic learning rates of DeiT and Swin are 1e-3 and 5e-4, and the weight decay of 0.05, like OpenMixup settings, we also used some RandAugment~\cite{cubuk2020randaugment} operations. We reported the best top-1 accuracy in the last 10 training epochs, and the results show that our SUMix also can bring effectiveness to ViTs.

Table \ref{tab: vit} shows the results of the ViT models using SUMix, which improves the classification results by an average of \gbf{0.57}\% and \gbf{0.23}\% at DeiT-Small and Swin-Tiny.

\paragraph{\textbf{Classification based on Saliency methods.}} To further explore the capability of SUMix, we not only apply it to some Hand-crafted methods, we apply SUMix to PuzzleMix and AutoMix, which are based on saliency-learning methods, and Tab \ref{tab: saliency} shows that SUMix will also bring some boosts, PuzzleMix and AutoMix has gained \gbf{0.30}\%, \gbf{0.26}\% at ResNet18, and \gbf{0.91}\%, \gbf{0.98}\% at ResNeXt50. Those methods combined with SUMix can even reach almost close to the performance of the current state-of-the-art method AdAutoMix.
\begin{figure}[t]
    \centering
    \includegraphics[scale=0.22]{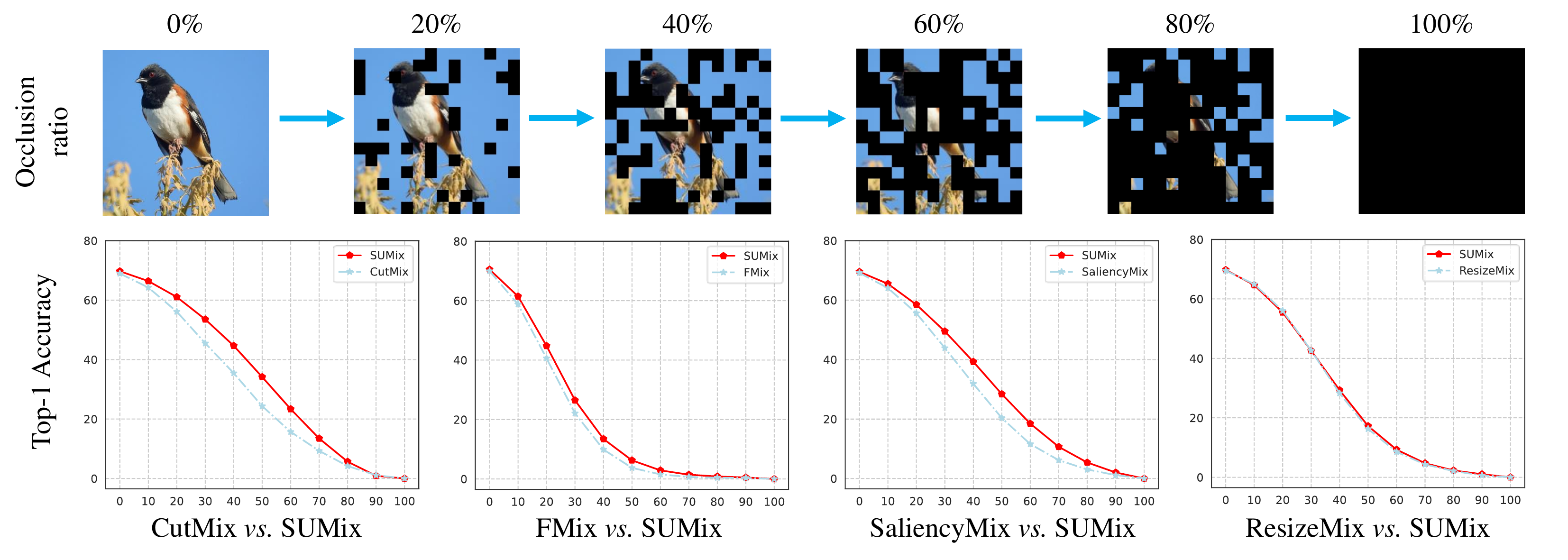}
    \caption{The top of the figure shows a visualization of the sample at 0\% to 100\% occlusion ratio. The light blue curves in the lower four subfigures show the classification accuracy of CutMix, FMix, SaliencyMix, and ResizeMix on ImageNet-1K using ResNet18 for 100 epochs of training, respectively, and the red curve shows the classification accuracy of these methods using SUMix.}
    \label{fig:5}
\end{figure}
\subsection{Occlusion Robustness}
To analyze the enhancement brought by SUMix to the robustness of the mixup methods, we randomly conducted masking patches experiment~\cite{naseer2021occlusion}. The experiment involved dividing the image into 16$\times$16 patches and masking them according to different ratios. The resulting image was then classified by the model. We used the ImageNet dataset, which contains 1000 daily life objects. It is more representative. The ResNet18 classifier was used to classify the masked images and verify the model's robustness capability. The curves at the bottom of Figure.\ref{fig:5} show that SUMix provides a performance gain compared to the corresponding mixup method at different occlusion ratios.

\subsection{Robustness}
We first evaluated robustness against corruption on CIFAR-100-C~\cite{hendrycks2019benchmarking}, which was designed manually to provide 19 different corruptions (\emph{e.g.}, noise, blur, fog, \emph{etc.}) to evaluate robustness. In Table \ref{tab:corr}, we show the results of improving some mixup methods with SUMix. We further investigate the FGSM~\cite{iclr2015fgsm} white box attack experiments with 8/255 $\ell_{\infty}$ epsilon ball and similarly show the results of the improvement of our approach on mixup methods in Table \ref{tab:corr}.
\begin{figure*}[t]
    \begin{minipage}{0.54\linewidth}
        \input{tabs/exp_corruptions_fgsm}
    \end{minipage}
    \begin{minipage}{0.46\linewidth}
        \input{tabs/exp_ablation}
    \end{minipage}
\end{figure*}
\subsection{Ablation Study}
In the ablation study, we primarily analyze the effect of the two SUMix modules in the model. The following two analyses are primarily done. (1) \textbf{Can redefining $\lambda$ bring gains in models?} (2) \textbf{Can uncertainty lead to regularization gains?} Table \ref{tab:abaltion} shows the experimental results for CutMix, SaliencyMix, and FMix using ResNet18 and ResNeXt50 on the CIFAR100 and CUB200 datasets.
\begin{enumerate}[(1)]
\setlength\topsep{0.5em}
\setlength\itemsep{0.0em}
\setlength\leftmargin{0.5em}
    \item Table \ref{tab:abaltion} shows that with our redefined  $\widetilde{\lambda}$, CutMix and SaliencyMix are improved by \gbf{0.96}\% and \gbf{0.67}\% on the CIFAR100 dataset, and FMix and SaliencyMix are improved by \gbf{1.75}\% and \gbf{0.69}\% on the CUB200 dataset, which means that the module can bring a significant increase in the classification ability, in other words, the redefined $\widetilde{\lambda}$ can obtain a more sensible loss function for the model to train.
    \item To calculate the regularization loss, we consider three cases: 1) using only the semantic information $\ell_s$, 2) using only the uncertainty information $\ell_u$, and 3) using both the semantic and uncertainty information $\ell_{su}$. Table \ref{tab:abaltion} shows that on the CUB200 dataset, FMix and SaliencyMix improve accuracy by 1.75\% and 0.57\% respectively when using only $\ell_s$, and by 1.96\% and 0.95\% respectively when using only $\ell_u$. When both types of information are used, accuracy improves by \gbf{1.75}\% and \gbf{1.21}\%. Similarly, on the CIFAR100 dataset, there was a \gbf{1.61}\% and \gbf{0.79}\% improvement. The experimental results show that combining semantic and uncertainty information can significantly improve accuracy.
\end{enumerate}

%% file: tabs/exp_generic.tex
\begin{table}[b]
    \centering
    \setlength{\tabcolsep}{1.5mm}
    \caption{Top-1 accuracy(\%)$\uparrow$ of mixup methods on CIFAR-100, Tiny-ImageNet and ImageNet-1K. * denotes mixup methods with SUMix.}
    \resizebox{1.0\linewidth}{!}{
        \begin{tabular}{l|ccc|cc|cc} 
        \toprule
            &\multicolumn{3}{c|}{CIFAR100}  &\multicolumn{2}{c|}{Tiny-ImageNet}  &\multicolumn{1}{c}{ImageNet-1K} \\
            Method       & ResNet18   & ResNeXt50  & W-ResNet28-8    & ResNet18   & ResNeXt50   & ResNet18   \\ \hline 
            CutMix       & 78.17      & 78.32      & 84.45           & 65.53      & 66.47       & 68.95   \\
            FMix         & 79.69      & 79.02      & 84.21           & 63.47      & 65.08       & 69.96  \\
            SaliencyMix  & 79.12      & 78.77      & 84.35           & 64.60      & 66.55       & 69.16  \\
            ResizeMix    & 80.01      & 80.35      & 84.87           & 63.74      & 65.87       & 69.50  \\ \hline
            \rowcolor{mygray} CutMix$^*$  & \textbf{79.78} & \textbf{79.91} & \textbf{84.56}  & \textbf{65.71} & \textbf{68.74}    & \textbf{69.71} \\
            \rowcolor{mygray} FMix$^*$    & \textbf{80.20} & \textbf{80.79} & \textbf{84.32}  & \textbf{63.69} & \textbf{67.12}      & \textbf{70.48} \\
            \rowcolor{mygray} SaliencyMix$^*$  & \textbf{79.91} & \textbf{79.32} & \textbf{84.58}      & \textbf{65.68} & \textbf{68.92}      & \textbf{69.52} \\
            \rowcolor{mygray} ResizeMix$^*$    & \textbf{80.38} & \textbf{80.72} & \textbf{84.91}      & \textbf{65.30} & \textbf{67.49}      & \textbf{69.76} \\ \hline
            Avg. Gain   & \gbf{+0.82}   & \gbf{+1.07}   & \gbf{+0.12}       & \gbf{+0.81}   & \gbf{+2.07}    & \gbf{+0.47}   \\
            \bottomrule
        \end{tabular}
}
    \label{tab:gener}
\end{table}

%% file: tabs/exp_fine_grained.tex
\begin{table}[t]
    \centering
    \setlength{\tabcolsep}{2.5mm}
    \caption{Top-1 accuracy(\%)$\uparrow$ of mixup methods on CUB200, FGVC-Aircrafts. * denotes mixup methods SUMix. $\dag$ denotes that we reproduce the results of that experiment.}
    \resizebox{0.75\linewidth}{!}{
        \begin{tabular}{l|cc|cc} 
        \toprule
                         &\multicolumn{2}{c|}{CUB200}             & \multicolumn{2}{c}{FGVC-Aircrafts}        \\
            Method       & ResNet18   & ResNeXt50   & ResNet18   & ResNeXt50    \\ \hline       
            CutMix$^\dag$      & 77.70 & 83.67     & 78.84 & 84.55   \\
            FMix        & 77.28 & 84.06     & 79.36 & 84.10     \\
            SaliencyMix$^\dag$ & 75.77 & 82.83     & 79.78 & 84.31    \\
            ResizeMix   & 78.50 & 84.16     & 78.10 & 84.08    \\ \hline
            \rowcolor{mygray} CutMix$^*$       & \textbf{78.20} & \textbf{83.71}       & \textbf{79.72} & \textbf{85.84}   \\
            \rowcolor{mygray} FMix$^*$         & \textbf{79.24} & \textbf{84.33}       & \textbf{79.48} & \textbf{84.64}   \\
            \rowcolor{mygray} SaliencyMix$^*$  & \textbf{76.98} & \textbf{82.84}       & \textbf{79.90} & \textbf{84.49}   \\
            \rowcolor{mygray} ResizeMix$^*$    & \textbf{78.56} & \textbf{84.23} & \textbf{80.29}      & \textbf{85.12}    \\ \hline
            Avg. Gain        & \gbf{+0.93}   & \gbf{+0.10}  & \gbf{+0.82}    & \gbf{+0.67}         \\
            \bottomrule
        \end{tabular}
}
    \label{tab: fine}
    \vspace{-1.0em}
\end{table}

%% file: tabs/vit_cls.tex
\begin{table}[H]
    \centering
    \setlength{\tabcolsep}{1.75mm}
    \caption{Top-1 accuracy(\%)$\uparrow$ of mixup methods on CIFAR100. * denotes mixup methods with SUMix.}
    \resizebox{1.0\linewidth}{!}{
        \begin{tabular}{l|cccc} 
        \toprule
                                               &\multicolumn{4}{c}{CIFAR100}        \\
            Method                             & \multicolumn{2}{c}{DeiT-Small}     & \multicolumn{2}{c}{Swin-Tiny}         \\ \hline       
            CutMix                             & 74.12  &          & 80.64  &       \\
            FMix                               & 70.41  &          & 80.72  &       \\
            SaliencyMix                        & 69.78  &          & 80.40  &       \\
            ResizeMix                          & 68.45  &          & 80.16  &       \\ \hline
            \rowcolor{mygray} CutMix$^*$       & \textbf{75.26} & \gbf{+1.14} & \textbf{80.83} & \gbf{+0.19}    \\
            \rowcolor{mygray} FMix$^*$         & \textbf{70.69} & \gbf{+0.28} & \textbf{80.73} & \gbf{+0.01}    \\
            \rowcolor{mygray} SaliencyMix$^*$  & \textbf{70.31} & \gbf{+0.53} & \textbf{80.71} & \gbf{+0.29}    \\
            \rowcolor{mygray} ResizeMix$^*$    & \textbf{68.78} & \gbf{+0.33}     & \textbf{80.59} & \gbf{+0.43}    \\
            \bottomrule
        \end{tabular}
}
    \label{tab: vit}
\end{table}

%% file: tabs/saliency_mixup_cls.tex
\begin{table}[H]
    \centering
    \setlength{\tabcolsep}{1.75mm}
    \caption{Top-1 accuracy(\%)$\uparrow$ of saliency-based mixup methods on CIFAR100. * denotes mixup methods with SUMix.}
    \resizebox{1.0\linewidth}{!}{
        \begin{tabular}{l|cccc} 
        \toprule
                                               &\multicolumn{4}{c}{CIFAR100}        \\
            Method                             &\multicolumn{2}{c}{ResNet18}     & \multicolumn{2}{c}{ResNeXt50}       \\ \hline       
            PuzzleMix                          & 81.13    &     & 81.69     &       \\
            AutoMix                            & 82.04    &     & 82.84     &       \\ \hline
            \rowcolor{mygray} PuzzleMix$^*$    & \textbf{81.43} & \gbf{+0.30} & \textbf{82.60} & \gbf{+0.91}   \\
            \rowcolor{mygray} AutoMix$^*$      & \textbf{82.30} & \gbf{+0.26} & \textbf{83.82} & \gbf{+0.98}   \\
            \rowcolor{pp}AdAutoMix                          & 82.32    &     & 83.81     &       \\ 
            \bottomrule
        \end{tabular}
}
    \label{tab: saliency}
\end{table}

%% file: tabs/exp_corruptions_fgsm.tex
\begin{table}[H]
    \centering
    \setlength{\tabcolsep}{1.5mm}
    \caption{Top-1 acc(\%)$\uparrow$ and FGSM error(\%)$\downarrow$ of ResNet18 without and with SUMix.}
    \resizebox{1.0\linewidth}{!}{
        \begin{tabular}{l|cc|cc|cc}
        \toprule
                         &\multicolumn{2}{c|}{Clean} & \multicolumn{2}{c|}{Corruption} &\multicolumn{2}{c}{FGSM} \\
                         & \multicolumn{2}{c|}{Acc(\%)$\uparrow$} &\multicolumn{2}{c|}{Acc(\%)$\uparrow$} &\multicolumn{2}{c}{Error(\%)$\downarrow$} \\ \hline
            \rowcolor{mygray} Method       & MCE   & \bf{SUMix}       & MCE   & \bf{SUMix}       & MCE   & \bf{SUMix} \\ \hline       
            CutMix       & 78.17 & \bf{79.78}     & 43.06 & \bf{44.31}     & 91.15 & 90.41 \\
            FMix         & 79.69 & \bf{80.20}     & 48.79 & \bf{49.14}     & 89.16 & 89.08 \\
            SaliencyMix  & 79.12 & \bf{79.91}     & 43.73 & \bf{44.36}     & 89.64 & \textcolor{gray}{91.49} \\
            ResizeMix    & 80.01 & \bf{80.38}     & 46.12 & \bf{46.28}     & 90.04 & \textcolor{gray}{91.05} \\
            \bottomrule
        \end{tabular}
}
    \label{tab:corr}
\end{table}

%% file: tabs/exp_ablation.tex
\begin{table}[H]
    \centering
    \setlength{\tabcolsep}{1.5mm}
    \caption{Ablation experiments on CIFAR100, CUB200 based on ResNet18.}
    \resizebox{1.0\linewidth}{!}{
        \begin{tabular}{l|cc|cc}
        \toprule
                                     &\multicolumn{2}{c|}{CIFAR100}  &\multicolumn{2}{c}{CUB200}\\ 
        Method                       & CutMix    & SaliencyMix       & FMix    & SaliencyMix \\ \hline
        baseline & \textcolor{gray}{78.17} & \textcolor{gray}{79.12} & \textcolor{gray}{77.28} & \textcolor{gray}{75.77} \\
        $+\widetilde{\lambda}$                 & 79.13     & 79.79             & 79.03     & 76.46 \\
        $+\ell_s$                       & 79.51     & 78.71             & 79.03     & 76.34 \\
        $+\ell_u$                       & 79.60     & 79.43             & 79.24     & 76.72 \\
        \rowcolor{mygray} $+\ell_{su}$  & \bf{79.78}& \bf{79.91}  & \bf{79.24} & \bf{76.98} \\
        \bottomrule
        \end{tabular}
}
    \label{tab:abaltion}
\end{table}

%% file: 6_Conclusion.tex
\section{Conclusion and Limitations}
\label{sec: conclusion & limitations}

\paragraph{\textbf{Conclusion.}} In this paper, we propose the SUMix approach to regularize the DNN in terms of the semantic information and uncertainty between the raw and mixed samples. Specifically, the semantic information difference between the raw and mixed samples is used to obtain the mix ratio $\lambda$, while the uncertainty is, in turn, combined with the semantic information to obtain a regularized loss function. Experimental results on five datasets proved the effectiveness of SUMix for the regularization of the mixup methods.

\paragraph{\textbf{Limitations.}} Currently, SUMix is due to learn how to obtain a sensible mixing ratio $\lambda$ and an uncertainty-based feature vector. These are currently proven only on classification tasks. In future work, we would like to apply and improve SUMix accordingly on some downstream tasks \emph{e.g.} segmentation, detection, and other methods. And for semi-supervised tasks, we can get more accurate pseudo-labels for training models.

%% file: Appendix.tex
\section*{Appendix A}

\subsection*{A.1 More Information about $\zeta$}
\label{appendix:a1}
We reported the hyper-parameter $\zeta$ of $\mathcal{L}_{su}$ on the CIFAR100 dataset shown in Table \ref{tab:zeta}. And for the fine-grained dataset, CUB200 and FGVC-Aircrafted. We reported the settings in Table \ref{tab:zeta_cub} and Table \ref{tab:zeta_fgvc}.
\input{tabs/zeta_CIFAR100}
\begin{figure*}[h]
    \begin{minipage}{0.5\linewidth}
        \input{tabs/zeta_CUB200}
    \end{minipage}
    \begin{minipage}{0.5\linewidth}
        \input{tabs/zeta_FGVC}
    \end{minipage}
\end{figure*}

\subsection*{A.2 More Visualization Comparison}
\label{appendix:a2}
\begin{figure}[H]
    \centering
    \includegraphics[scale=0.30]{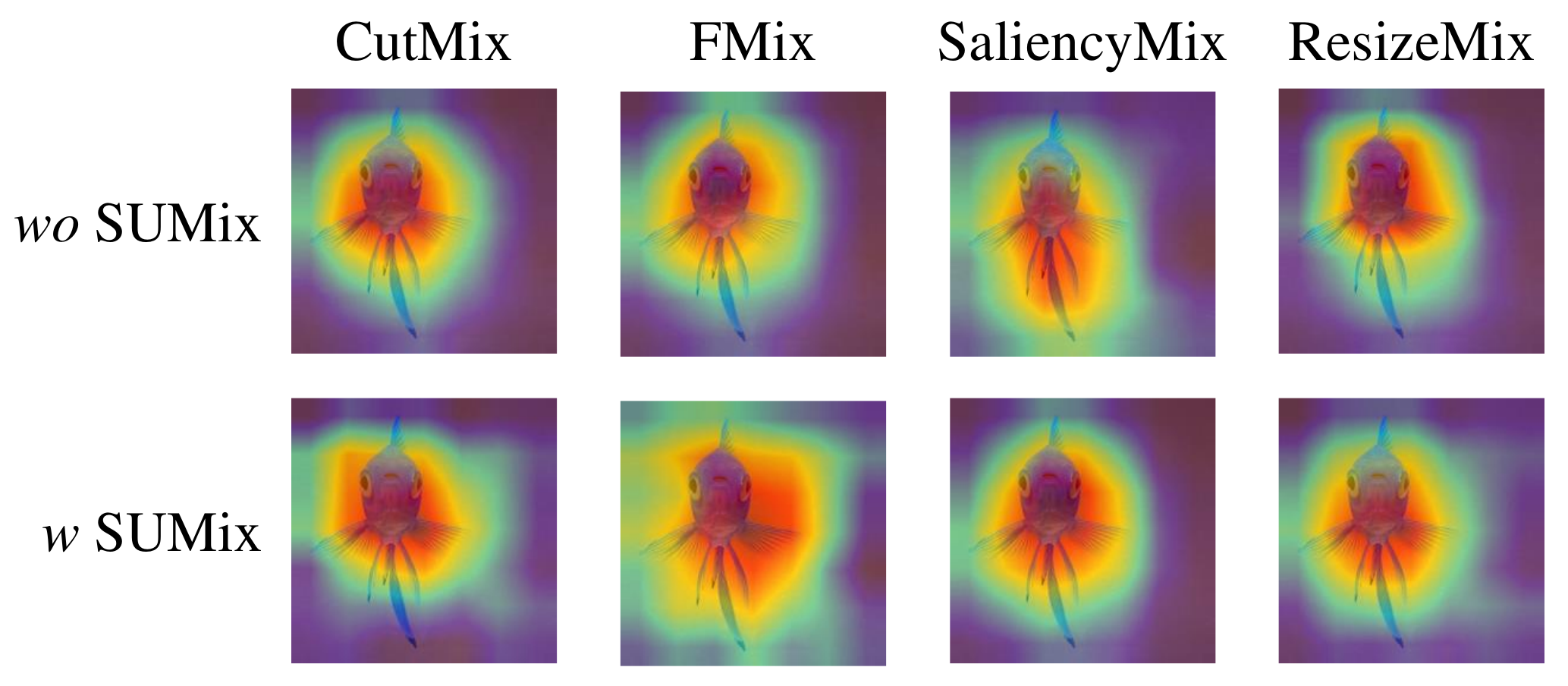}
    \caption{CAM visualization of the ResNet18 with and without SUMix trained on the ImageNet-1K dataset.}
    \label{fig:cam1}
\end{figure}

\begin{figure}[t]
    \centering
    \includegraphics[scale=0.32]{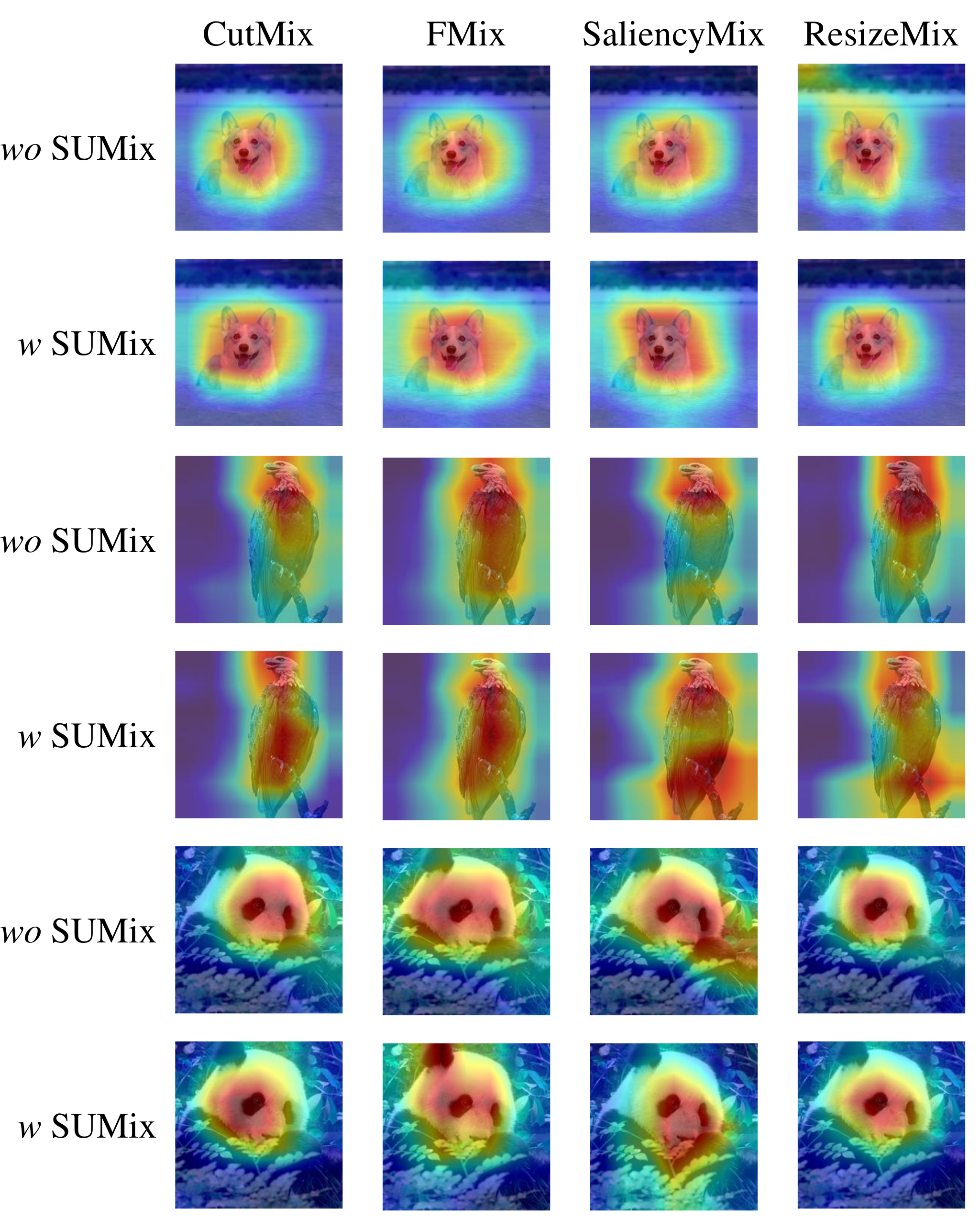}
    \caption{CAM visualization of the ResNet18 with and without SUMix trained on the ImageNet-1K dataset.}
    \label{fig:cam2}
\end{figure}

In Figure.\ref{fig:cam1} and Figure.\ref{fig:cam2}, we show the visualization obtained by the mixup method using the CAM method with and without SUMix.

%% file: tabs/zeta_CIFAR100.tex
\begin{table}[H]
    \vspace{-20pt}
    \centering
    \setlength{\tabcolsep}{1.5mm}
    \caption{The $\zeta$ setting on the CIFAR100 dataset.}
    \resizebox{1.0\linewidth}{!}{
        \begin{tabular}{l|ccccc} 
        \toprule
                           &\multicolumn{5}{c}{CIFAR100} \\
            $\zeta$ setting      & ResNet18   & ResNeXt50  & W-ResNet28-8    & DeiT-Small   & Swin-Tiny    \\ \hline 
            CutMix        & 0.5        & 0.5        & 0.5             & 0.5          & 0.5        \\
            FMix          & 1.0        & 1.0        & 0.5             & 0.5          & 0.5        \\
            SaliencyMix   & 0.5        & 1.0        & 0.5             & 0.5          & 0.5        \\
            ResizeMix     & 0.2        & 0.5        & 0.5             & 0.5          & 0.5        \\ 
            \bottomrule
        \end{tabular}
}
    \label{tab:zeta}
    \vspace{-60pt}
\end{table}

%% file: tabs/zeta_CUB200.tex
\begin{table}[H]
    \centering
    \setlength{\tabcolsep}{1.5mm}
    \caption{The $\zeta$ setting on the CUB200 dataset.}
    \resizebox{1.0\linewidth}{!}{
        \begin{tabular}{l|cc} 
        \toprule
                           &\multicolumn{2}{c}{CUB200} \\
            $\zeta$ setting       & ResNet18   & ResNeXt50     \\ \hline 
            CutMix        & 1.0        & 0.5           \\
            FMix          & 0.5        & 0.1           \\
            SaliencyMix   & 0.5        & 0.2           \\
            ResizeMix     & 0.5        & 0.5           \\ 
            \bottomrule
        \end{tabular}
}
    \label{tab:zeta_cub}
\end{table}

%% file: tabs/zeta_FGVC.tex
\begin{table}[H]
    \centering
    \setlength{\tabcolsep}{1.5mm}
    \caption{The $\zeta$ setting on the FGVC-Aircrafts dataset.}
    \resizebox{1.0\linewidth}{!}{
        \begin{tabular}{l|cc} 
        \toprule
                           &\multicolumn{2}{c}{FGVC-Aircrafts} \\
            $\zeta$ setting       & ResNet18   & ResNeXt50     \\ \hline 
            CutMix        & 1.0        & 0.5           \\
            FMix          & 0.5        & 0.5           \\
            SaliencyMix   & 1.0        & 0.5           \\
            ResizeMix     & 0.5        & 0.5           \\ 
            \bottomrule
        \end{tabular}
}
    \label{tab:zeta_fgvc}
\end{table}